\pdfoutput=1
\documentclass[11pt]{article}

\usepackage[final]{acl}

\usepackage{times}
\usepackage{latexsym}
\usepackage{hyperref}
\usepackage{url}
\usepackage{graphicx} 
\usepackage{amsthm}
\usepackage{footmisc}
\usepackage{wrapfig}

\usepackage{booktabs}
\usepackage{multirow}
\usepackage[table]{xcolor}
\usepackage[normalem]{ulem}
\usepackage{graphicx}
\usepackage{threeparttable}
\usepackage{colortbl}
\usepackage{xcolor}
\usepackage[most]{tcolorbox}
\usepackage{listings} 
\usepackage{longtable}
\usepackage{tablefootnote}
\usepackage{caption}
\usepackage{amsmath}
\usepackage{amsfonts}
\usepackage{amssymb}

\usepackage[T1]{fontenc}
\usepackage[utf8]{inputenc}

\usepackage{microtype}

\title{Mock Worlds, Real Skills: Building Small Agentic Language Models with Synthetic Tasks, Simulated Environments, and Rubric-Based Rewards}

\author{%
Yuanjie Lyu$^{1}$,
Chengyu Wang$^{2}$\thanks{Corresponding author.},
Lei Shen$^{3}$,
Jun Huang$^{2}$,
Tong Xu$^{1}$\footnotemark[1]\\
$^{1}$ University of Science and Technology of China\\
$^{2}$ Researcher\\
$^{3}$ Xi'an Jiaotong University\\
\texttt{s1583050085@gmail.com}%
}

\begin{document}

\maketitle

\begin{abstract}
Small LLMs often struggle to match the agentic capabilities of large, costly models. While reinforcement learning can help, progress has been limited by two structural bottlenecks: existing open-source agentic training data are narrow in task variety and easily solved; real-world APIs lack diversity and are unstable for large-scale reinforcement learning rollout processes. We address these challenges with \textsc{SynthAgent}, a framework that jointly synthesizes diverse tool-use training data and simulates complete environments. Specifically, a strong teacher model creates novel tasks and tool ecosystems, then rewrites them into intentionally underspecified instructions. This compels agents to actively query users for missing details. When handling synthetic tasks, an LLM-based user simulator provides user-private information, while a mock tool system delivers stable tool responses. For rewards, task-level rubrics are constructed based on required subgoals, user-agent interactions, and forbidden behaviors. Across 14 challenging datasets in math, search, and tool use, models trained on our synthetic data achieve substantial gains, with small models outperforming larger baselines. \footnote{Code for data synthesis pipeline and training:~\url{https://github.com/haruhi-sudo/SYNTHAGENT}}
\end{abstract}

\section{Introduction}
\begin{figure}[t]
\centering
\includegraphics[width=0.49\textwidth]{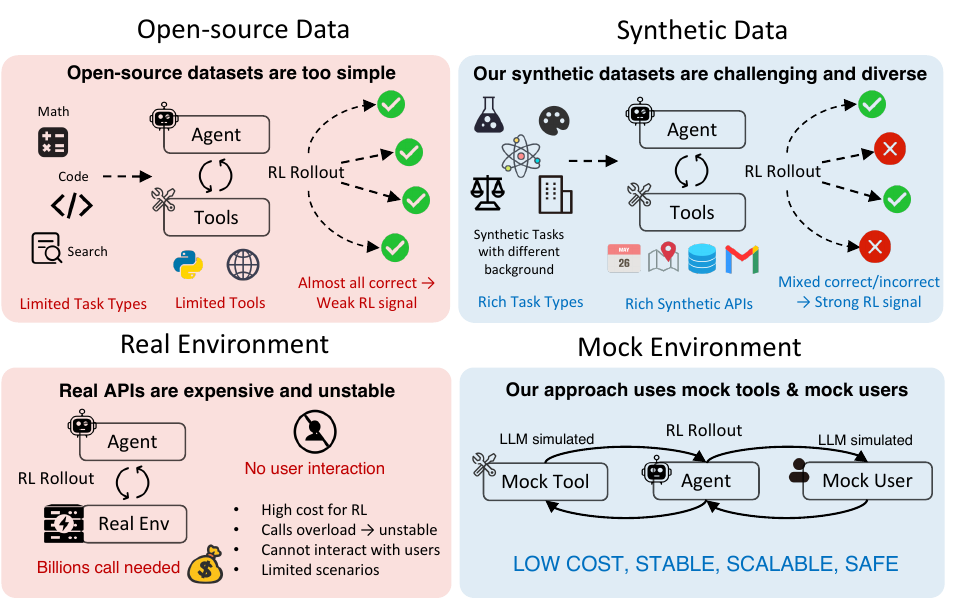}
\caption{
Comparison between existing agentic RL training recipes and ours. Open-source agentic training data are narrow in domain, while real-world APIs are costly and unstable. We replace these with diverse synthetic tasks and associated mock environments.
}
\label{fig:demo}
\end{figure}

Large language models (LLMs) demonstrate strong agentic capabilities within ReAct-style frameworks~\cite{yao2023react}. Through an iterative \emph{reasoning--action--observation} loop, LLM-based agents can solve complex tasks that require interaction with external environments~\cite{xi2025rise}, such as booking hotels or canceling flights~\cite{barres2025tau2}. However, these agentic capabilities depend heavily on very large base models~\cite{team2025kimi}, resulting in substantial inference costs and deployment overhead. Consequently, enabling smaller models to reproduce the agentic capabilities of large models has become an important research direction~\cite{lyu2025correction,li2025chain}.

Distillation methods based on supervised fine-tuning (SFT), in which a student model clones a teacher's behavior~\cite{torabi2018behavioral}, can enhance the agentic capabilities of small models. Recent studies~\cite{mai2025agentrlscalinglaw} further show that reinforcement learning (RL) is more effective than SFT for improving long-horizon planning and adaptive decision-making. However, most RL-based approaches focus on refining RL algorithms themselves~\cite{dong2025agentic-entropy,dong2025agentic}, while overlooking two fundamental bottlenecks:
\begin{itemize}
\item \emph{Lack of diverse and challenging agentic training data.} Public datasets cover only a narrow range of domains and tools, and many have already been seen by modern LLMs during pre-training or fine-tuning. As a result, RL rollout often yields near-perfect trajectories with weak learning signals~\cite{yu2025dapo}.
\item \emph{Absence of stable, diverse environments.} Real environments rarely support real-time model-user interaction and offer only a narrow tool set. RL rollout also requires a massive number of tool calls, making it impractical to rely on costly real-world APIs~\cite{team2025longcat}.

% and RL rollout requires a massive number of tool calls, making it impractical to rely on costly and unstable real-world APIs~\cite{team2025longcat}.
\end{itemize}

To address these, we introduce \textsc{SynthAgent}, a framework that synthesizes tool-use tasks along with lightweight mock tool interfaces. A strong agentic teacher LLM generates novel tasks and their associated tools, guided by diverse persona backgrounds~\cite{ge2025scalingsyntheticdatacreation}. As shown in Figure~\ref{fig:demo}, each task is paired with its own tool ecosystem, greatly expanding task and tool diversity. Moreover, the synthetic tools require no real deployment: an open-source LLM simulates both user and tool responses locally, ensuring stability.

Specifically, \emph{for synthesizing training data}, we introduce an information gap by rewriting detailed workflows as underspecified instructions, while critical details are hidden in a private user context. This design forces agents to actively query users and call tools to recover missing information, encouraging genuine long-horizon interaction.
Second, \emph{for LLM-based tool response consistency}, we maintain a task-level mapping of prior tool calls and responses. New calls are answered by consulting this mapping for consistent replies. As each synthetic task has a unique toolset, the mapping is scoped per task, keeping it lightweight during rollout.
Finally, \emph{for reward design}, we avoid subjective LLM-written rubrics and derive rewards from observable behavior. Using the workflow from data synthesis as a reference, we extract corresponding high-level subgoals from real execution trajectories, each reachable via multiple valid paths. This yields execution-grounded rewards that support diverse strategies, while filtering out low-quality data when the teacher fails to reliably complete the workflow.

We evaluate our approach on 14 recent, challenging datasets spanning agentic tool use~\cite{yehudai2025survey} and short-horizon reasoning. In real-world tasks, models trained on synthetic data within virtual environments substantially outperform those trained on open-source datasets. After training, our 8B--14B models surpass a 32B model on multiple agentic benchmarks. In summary, the major contributions of this work are as follows:
\begin{itemize}
    \item We introduce an open-source framework for synthesizing diverse agentic tool-use tasks, with stable, lightweight mock tool interfaces.
    \item In synthetic tasks that require genuine long-horizon interactions, we train models with execution‑grounded, rubric‑based rewards.
    \item Extensive experiments on 14 challenging datasets demonstrate that models trained on synthetic data and virtual environments achieve strong real-world performance.
\end{itemize}

\section{Related Work}
\begin{figure*}[t]
\centering
\includegraphics[width=.99\textwidth]{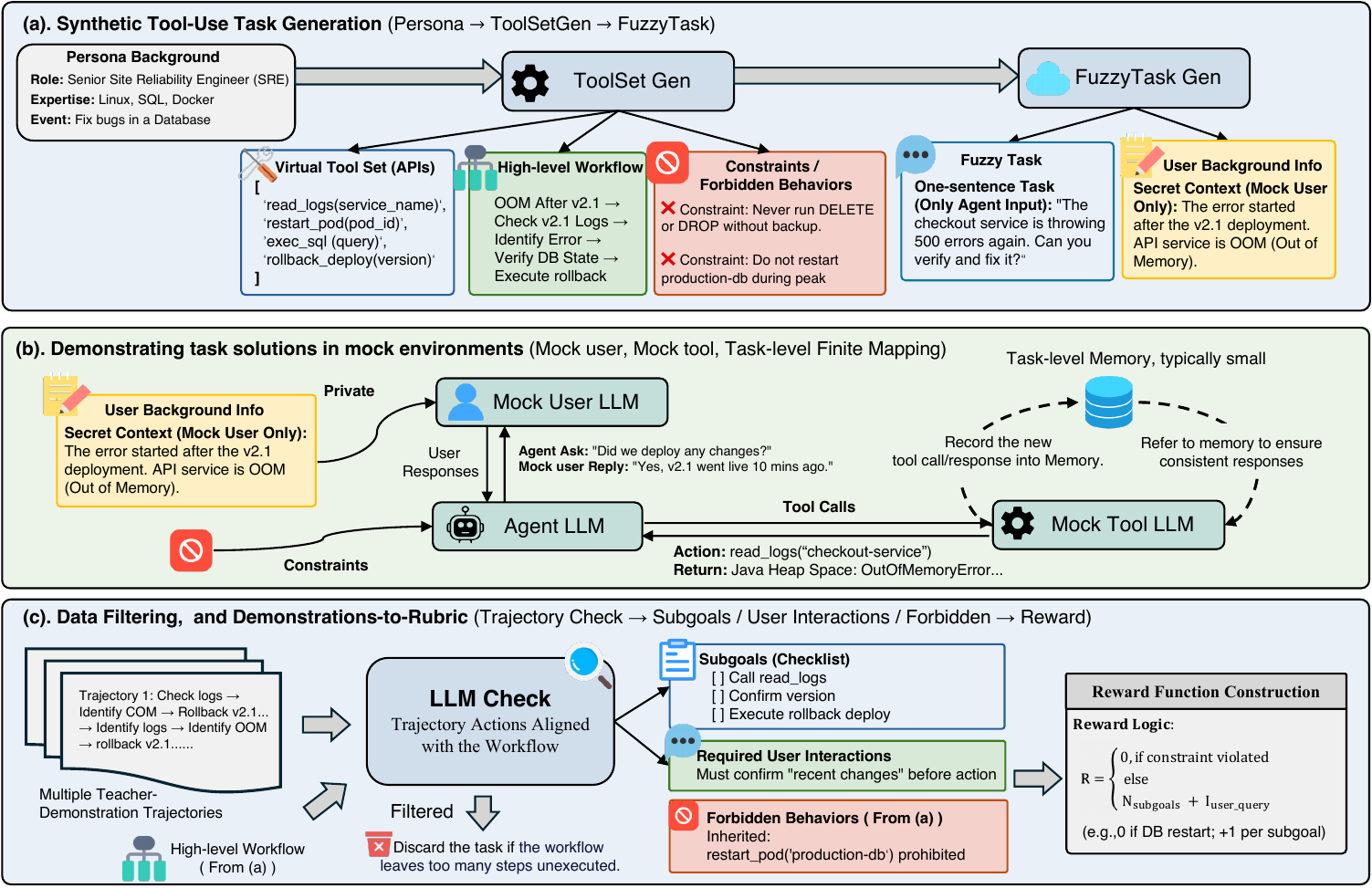}
\caption{
A unified pipeline for generating synthetic tool-use tasks, constructing stable mock environments, and deriving rubric-based rewards for agentic RL. Diverse tasks and tool ecosystems are created, guided by personas. For each synthetic task, an LLM-simulated user and environment are employed. To assign rewards, multiple trajectories are compared to the previously generated high-level workflow to infer task-specific rubrics.
}
\label{fig:method}
\end{figure*}

\subsection{Agentic Reinforcement Learning}

Recent studies show that RL outperforms SFT in long-horizon planning and adaptive decision-making~\cite{zhang2025landscape}, making RL a core paradigm for training LLM agents in dynamic, multi-turn environments~\cite{mialon2023gaia}. Classical methods such as Q-learning~\cite{mnih2015human}, PPO~\cite{schulman2017proximal}, and self-play~\cite{silver2017mastering} have provided the conceptual foundation for agentic optimization in LLM-based systems. These techniques have evolved into language-centric RL frameworks, where natural-language reasoning steps, tool calls, and observations are treated as latent states and actions~\cite{yao2023react,zhang2025landscape}. Recent work has further improved RL algorithms to better couple exploration with robust tool use in long-horizon tasks, including verifiable-reward RL~\cite{su2025crossing}, entropy-regularized policy optimization~\cite{dong2025agentic-entropy}, and agent-specific PPO/GRPO variants~\cite{dong2025agentic}. Despite this progress, research remains largely focused on RL algorithms, with considerably less attention given to data and environment design.
% Subsequent work introduced structured reasoning-action loops~\cite{yao2023react}, enabling LLMs to integrate chain-of-thought planning with environment interactions. 

\subsection{Synthetic Data for Agentic Training}

% A parallel line of research focuses on synthetic data generation and mock environments to overcome the instability and cost of real-world agent training. Synthetic trajectories produced by stronger teacher models allow scalable creation of diverse tasks and reasoning patterns~(Ge et al., 2025; SYNTHLLM, 2025). Persona-driven synthesis further expands behavioral diversity by conditioning generation on heterogeneous roles and preferences, yielding richer task distributions than standard instruction datasets. Scaling studies show that such synthetic corpora can support large-scale pretraining and RL, provided that diversity is maintained.

% To support RL rollouts, researchers have increasingly adopted LLM-based simulators and virtual tools. Systems like StableToolBench and Simia-RL introduce virtual API servers and LLM-driven world models, respectively, allowing agents to act in deterministic, high-throughput environments without the brittleness of live systems. These mock environments enable reproducible rewards, controllable failure modes, and rapid rollout generation—properties essential for stable RL training.

The effectiveness of agentic RL depends on high-quality, diverse data and environments, which remain scarce~\cite{yehudai2025survey}. Early works such as Self-Instruct~\citep{wang2023self} use strong but closed-source LLMs to generate instruction-following data for training smaller open-source models. To further increase diversity, \citet{ge2025scalingsyntheticdatacreation} propose Persona Hub, which curates one billion web-derived personas to enable diverse synthetic data generation. In parallel, \citet{qin2025scaling} identify a ``rectified scaling law'' for synthetic data: as long as diversity is maintained, gains from synthetic pre-training persist even at very large scales.

% In environment construction, benchmarks such as ToolBench~\cite{qin2023toolllm} rely on real-world APIs for authenticity, or use LLMs to simulate existing APIs to reduce cost (e.g., StableToolBench~\cite{guo2025stabletoolbench}; \citet{li2025simulating}). However, such simulations remain constrained by real back-end services, making it difficult to enhance environmental diversity while maintaining stability. 

For environment construction, benchmarks~\cite{qin2023toolllm} rely on real-world APIs for authenticity or use LLMs to simulate existing APIs and reduce cost. However, these environments remain constrained by real services, a lack of diversity~\cite{lu2025toolsandbox}. And identical states and actions can produce inconsistent responses in simulation, making them unsuitable for direct RL.

Most existing synthesis methods neither target long‑horizon agentic tasks nor construct stable RL environments, meaning few works approach ours on either front. Our approach further unifies both aspects and integrates them with rubric‑based RL.

% Our approach innovates on both, and combines them with rubric-based RL to enhance genuine long-horizon agentic abilities.

\section{Methodology}
% \begin{figure*}[t]
% \centering
% \includegraphics[width=.985\textwidth]{method.pdf}
% \caption{
% A unified pipeline for generating synthetic tool-use tasks, constructing stable mock environments, and deriving rubric-based rewards for agentic RL. We create diverse tasks and tool ecosystems guided by personas. When handing the synthetic task, an LLM-simulated user and environment are employed. For rewards, we compare multiple trajectories with the previously generated high-level workflow to infer task-specific rubrics.
% }
% \label{fig:method}
% \end{figure*}

In this section, we move beyond improving RL algorithms themselves, and instead focus on two fundamental yet underexplored factors that limit agentic RL for LLMs: diverse, challenging training data and diverse, stable environments. Figure~\ref{fig:method} illustrates \textsc{SynthAgent}, a framework for synthesizing tool-use tasks, constructing virtual environments, and deriving rubric-based task rewards.

% The synthetic task generation phase (Phase 1) produces diverse tasks along with their tool ecosystems. When the model deals with these tasks (Phase 2), an LLM-simulated user and a mock tool system form a stable virtual environment with consistent responses. Finally, the Demonstrations-to-Rubric phase (Phase 3) converts teacher trajectories into task-specific rubrics, which serve as reward signals during RL training of small agentic models.

% Each phase presents distinct challenges. Synthetic tasks may become over-specified and thus uninformative for RL. The virtual environment must provide consistent, low-variance tool behavior. Reward construction must avoid subjective, hand-crafted rules while remaining scalable across tasks. Our design addresses these challenges through coordinated components described below.

\subsection{Synthetic Tool-Use Task Generation}
\label{sec:tool_use_task_generation}

\paragraph{Tool Set Generation}

Existing open-source agentic datasets are predominantly composed of web search and math tasks, which typically can be solved using only a search tool or a code interpreter, resulting in homogeneous tool environments. To diversify settings and tool usage, we incorporate large-scale personas from Persona Hub~\cite{ge2025scalingsyntheticdatacreation} as backgrounds (Figure~\ref{fig:method}(a)). These personas cover a wide range of identities and scenarios (e.g., a senior SRE debugging a database issue).

For each persona-defined context, we employ a strong agentic LLM to (i) infer a high-level workflow describing how an individual with that background would accomplish the task, and (ii) based on this inferred workflow, construct a task-specific virtual tool ecosystem with tool descriptions and API specifications. As a result, each task is paired with a dedicated tool suite, encouraging models to learn tool-use procedures rather than memorize fixed APIs. To further increase task difficulty, we introduce task-level forbidden constraints, such as ``disallowing a system reboot during a database repair task''. These constraints raise the RL challenge by requiring the model to plan and act within nontrivial restrictions.\footnote{For data synthesis, we employ Qwen3-235B-A22B-Instruct-2507~\cite{yang2025qwen3} due to its strong agentic capabilities and low-cost local deployment.}

\paragraph{Fuzzy Task Generation} 

After defining the tool set available to the agent, we next design the tasks, i.e., the agent's inputs. Tasks derived directly from previously generated high-level workflows are often over-specified; for example, \emph{Check logs $\rightarrow$ Verify DB state $\rightarrow$ Execute rollback}. The initial input outlines the workflow, and simply following it becomes the optimal action sequence $a^\star$. Consequently, rollouts $\tau \sim \pi_\theta$ are highly homogeneous. At many steps $t$, the policy $\pi_\theta(a \mid s_t)$ is nearly deterministic; thus, during RL training, the variation of the advantage under $\pi_\theta$ may become negligible:
\begin{equation}
\mathrm{Var}_{a_t \sim \pi_\theta(\cdot \mid s_t)}\!\big[A(s_t,a_t)\big] \approx 0,
\label{eq:adv_var_collapse}
\end{equation}
where $A(s_t,a_t)$ is the advantage under a value baseline, satisfying $\mathbb{E}_{a_t \sim \pi_\theta(\cdot \mid s_t)}[A(s_t,a_t)] = 0$. As demonstrated by the Cauchy–Schwarz inequality:
\begin{equation}
\label{eq:pg_cs}
\begin{aligned}
\Big\| \mathbb{E}_{a_t \sim \pi_\theta(\cdot \mid s_t)} &\big[\nabla_\theta \log \pi_\theta(a_t \mid s_t)\, A(s_t,a_t)\big] \Big\| \\
\le\,&
\sqrt{\mathbb{E}_{a_t \sim \pi_\theta(\cdot \mid s_t)} \big[ \| \nabla_\theta \log \pi_\theta(a_t \mid s_t) \|^2 \big]} \\
&\cdot
\sqrt{\mathbb{E}_{a_t \sim \pi_\theta(\cdot \mid s_t)} \big[\| A(s_t,a_t) \|^2 \big]} \\
=\, &
\sqrt{\mathbb{E}_{a_t \sim \pi_\theta(\cdot \mid s_t)} \big[ \| \nabla_\theta \log \pi_\theta(a_t \mid s_t) \|^2 \big]} \\
&\cdot
\sqrt{\mathrm{Var}_{a_t \sim \pi_\theta(\cdot \mid s_t)} \big[A(s_t,a_t) \big]},
\end{aligned}
\end{equation}
when Eq.~(\ref{eq:adv_var_collapse}) holds, the expected gradient magnitude shrinks, weakening the learning signal.

To mitigate this degeneracy, we inject an \emph{information gap} during task construction: partition each task initial state $s_0$ into an agent-visible instruction $I$ and a user-only hidden context $C$:
\begin{equation}
s_0 \mapsto (I,C), \qquad \text{s.t. } H(a^\star \mid I) \gg \epsilon.
\end{equation}

Initially, $I$ is insufficient to determine the optimal action $a^\star$ uniquely; the conditional entropy $H(a^\star \mid I)$ is large. Critical details must be recovered through interaction. As illustrated in Figure~\ref{fig:method}(a), we employ an LLM to rewrite an overly explicit request into a minimal one $I$, for example, ``The checkout service is returning 500 errors again. Can you investigate and fix it?'' Decisive details (e.g., ``v2.1 was just deployed'' and ``OOM'') are moved to $C$ and revealed only when the agent queries the user.

Under intentionally high $H(a^\star \mid I)$, the policy must first query for the missing context $C$ before invoking tools. As observations $o_{\le t}$ gradually reveal the hidden information, uncertainty decreases ($H(a^\star \mid I, o_{\le t}) < H(a^\star \mid I)$). Early decisions therefore become nontrivial: the agent must decide which clarification to ask first to elicit informative observations $o_t$, preventing $\pi_\theta(\cdot \mid s_t)$ from becoming near-deterministic. This mitigates the gradient degeneration issue during model training.

Appendix~\ref{sec:example} provides examples of the synthetic tasks and tools.

\subsection{Mock Environments}
\label{sec:mock}
\paragraph{Mock Tool \& User} 
When the model attempts the synthetic tool-use tasks described above, the corresponding tool set is registered in the system prompt. Because these tools are virtual rather than real, we must simulate both tool execution and responses. To this end, we build a fully LLM-simulated mock environment, which requires no real deployment and supports large-scale interaction during RL rollout. The LLM-simulated tool receives the model's tool-call requests and returns appropriate outputs. The LLM-simulated user answers the model's queries based on user-private background information $C$ generated earlier.\footnote{These interactions are primarily simple, formatted question-answering tasks requiring no powerful model. Thus, we implement this component using Qwen3-30B-A3B-Instruct-2507~\cite{yang2025qwen3}, which is easy to deploy locally and incurs very low runtime cost.}

% Prior work~\cite{guo2025stabletoolbench} has explored using LLMs to simulate tools; we further introduce a user simulator that holds critical private information and ensures reproducible tool responses.

\paragraph{Stable Environments}  
% During the RL rollout, the same task may be attempted many times. This raises a central concern: identical tool calls can yield different responses across attempts, undermining training stability. 

% During the RL rollout, the same task may be attempted many times, raising a central concern: if tool response is non-reproducible, then even under the same state $s$, executing the same tool action $a$ (identical $\texttt{tool}+\texttt{args}$) may produce different observations $o$, injecting additional randomness into the transition distribution $P(o\mid s,a)$. This stochasticity propagates through the trajectory, increasing the variance of the trajectory-level score $R(\tau)$, which amplifies noise in advantage estimation, and ultimately yields a higher-variance gradient estimator $\nabla_\theta \log \pi_\theta(a_t\!\mid s_t)\,\hat A_t$, hindering training stability. 

During the RL rollout process, the same task is attempted many times, raising a central concern: if tool responses are non-reproducible, then even under the same state $s$, executing the same tool action $a$ (identical $\texttt{tool}+\texttt{args}$) may yield different observations $o$.  
This randomness propagates along the trajectory, so that the same $(s_t, a_t)$ can induce different future returns across rollouts, making the advantage estimate $\hat{A}_t$ inconsistent. Consequently, even for identical $(s_t, a_t)$, the policy update term $\hat{g}_t = \nabla_\theta \log \pi_\theta(a_t \mid s_t) \, \hat{A}_t$ may exhibit substantially different magnitudes and even opposite signs across samples, hindering training stability.

A natural mitigation is to add retrieval-augmented memory~\cite{lewis2020retrieval} to the tool simulator, storing past tool calls and responses. When generating a new tool response, the simulator retrieves similar calls as references to ensure consistency. In our setting, each task has its own tool suite, so only a few calls require within-task consistency. Instead of a full memory system, we use a lightweight task-level finite mapping:
\begin{equation}
\mathcal{M} = \{(u_i, y_i)\}_{i=1}^{M}, \quad u_i = (\texttt{tool}_i, \texttt{args}_i).
\end{equation}
When the model issues a valid tool call $u$, the simulator checks similar entries in $\mathcal{M}$ and checks for a semantically equivalent one. If none is found, it generates a response $y$ and adds $(u, y)$ to the mapping. Equivalence checking and response generation can be handled in a single forward pass, adding no extra computational cost.

Empirically, the task-level mapping remains very small. For example, in a rollout with 16 trajectories and an average of 10 tool calls per trajectory, even if all calls were unique, the size of $\mathcal{M}$ would be at most 160. Therefore, we can \emph{omit retrieval} altogether and include $\mathcal{M}$ directly in the tool simulator's prompt, allowing the model to identify matches. The entire process remains lightweight and efficient, significantly improves training stability without restricting exploration.

\subsection{Automatic Rubric-Based Rewards}

\paragraph{Task-Level Rubrics} 

% Unlike math tasks with clearly defined correctness-based rewards~\cite{guo2025deepseek}, reward design for multi-step, tool-use tasks is inherently challenging. A common practice is to use an LLM-as-a-judge to assign a scalar score to each trajectory, but such free-form judgments can be subjective. Fortunately, we construct fuzzy tasks by rewriting a high-level workflow, whose steps provide objective subgoals for trajectory reward.
Unlike math tasks with clearly defined correctness-based rewards~\cite{guo2025deepseek}, reward design for multi-step tool-use tasks is inherently challenging. A common practice is to use an LLM as a judge to assign a scalar score to each trajectory, but such judgments can be subjective. Fortunately, we construct fuzzy tasks by rewriting high-level workflows, whose steps provide objective subgoals (e.g., \emph{Check logs, Verify DB state, Execute rollback}) for trajectory rewards.

However, the designed workflow may not match real executions. To address this, we collect multiple actual executed trajectories from strong teacher models. Using the workflow as a reference, we prompt an LLM to extract workflow-relevant subgoals and user-agent interactions from each trajectory. Both are grounded in the workflow and can be achieved via multiple exploration paths.

% 想了下，这段还是不适合放在footnote，而且下标已经够多了
If the workflow leaves too many steps unexecuted, the example is removed. This filters noisy data and reduces reliance on the number of teacher demonstrations, since trajectories with many unexecuted steps can be discarded.

% This is a simple and well-defined semantic matching task.

% the subgoals and interactions used for evaluation are grounded in actual executions rather than merely hypothesized.

Additionally, during tool set generation (Section~\ref{sec:tool_use_task_generation}), each task is paired with its own set of forbidden behaviors (e.g., disallowing a system reboot). Combined with the subgoals and interaction requirements, these form a complete task-level rubric unique to each synthetic task. Appendix~\ref{sec:example} provides examples of the generated rubrics.

\paragraph{Rubric-based Reward}
During RL training, we use an LLM as a judge to assign rewards based on the task-level rubric. Specifically,
\begin{equation}
R(\tau) = \mathbb{I}(\tau) \cdot \bigl(N_{\text{subgoals}}(\tau) + I_{\text{user\_query}}(\tau)\bigr).
\end{equation}
Here, $\mathbb{I}(\tau) \in \{0,1\}$ is 0 if and only if $\tau$ violates any rubric-specified forbidden behavior (yielding zero reward), and 1 otherwise. $N_{\text{subgoals}}(\tau) \in [0,1]$ is the fraction of subgoals completed, and $I_{\text{user\_query}}(\tau) \in [0,1]$ is the fraction of required user–agent interactions satisfied; we average these scores as the final reward. This rubric-based reward design scales seamlessly to large numbers of synthesized tasks.

Table~\ref{tab:synthagent-data-stats-vertical} summarizes the statistics of our synthesized tool‑use dataset, and reports the total token cost of the full synthesis pipeline (including tool, task, and rubric generation). Since the entire process runs on locally deployed open-source models, the cost is negligible. And even with commercial APIs, it remains well within an affordable range.

%including the total number of tasks, the average number of interaction turns after training, the average size of the task-level mapping $|M|$, and the average number of tools defined per task.

\subsection{Final RL Training}
\label{sec:training}
% Following the technical reports of Kimi K2~\cite{team2025kimi} and LongCat~\cite{team2025longcat}, strong reasoning ability is essential for agentic tasks. Accordingly, we augment our virtual tool-use data with a small set of high-difficulty reasoning tasks, primarily challenging math problems. The reasoning data synthesis procedure is straightforward: we first select problems that small models consistently fail but strong models reliably solve. Since such problems are scarce, we then use a strong model to expand this set. Using persona information, each problem is rewritten into several scenario-based variants. Finally, we solve every synthesized problem with three models and retain only those with fully consistent answers to ensure reliability.

% Tool-use tasks run in a virtual environment with rubric-based rewards, while reasoning tasks run in a real Python environment and are evaluated by answer correctness. We then train the model on the combined dataset using GRPO~\cite{shao2024deepseekmath}. 

Following the technical reports of Kimi K2~\cite{team2025kimi} and LongCat~\cite{team2025longcat}, strong reasoning ability is essential for agentic tasks. Accordingly, we augment our virtual tool-use data with a small set of high-difficulty reasoning tasks, sampling 4,000 search or math instances from ToolStar~\cite{dong2025toolstarempoweringllmbrainedmultitool}.
Since our tool-use data contain rich contextual backgrounds whereas math problems are purely abstract, this mismatch may hinder training. To increase diversity, we prompt Qwen3-235B\footnote{For brevity, we use Qwen3-235B to represent Qwen3-235B-A22B-Instruct. The same applies in the following text.} with persona information~\cite{ge2025scalingsyntheticdatacreation} to rewrite each problem into a scenario-based variant. Each synthesized problem is then solved 3 times by Qwen3-235B, and we retain only those with fully consistent answers, ensuring reliability.

Tool-use tasks run in a virtual environment with rubric-based rewards, while reasoning tasks run in a real Python environment and are evaluated by answer correctness. We then train the model on the combined dataset using GRPO~\cite{shao2024deepseekmath}.

% Refer to Appendix~\ref{sec:implementation} for more details.

\section{Experiments}
\begin{table}[t]
\centering
\small
\setlength{\tabcolsep}{3pt}
\resizebox{0.485\textwidth}{!}{
\begin{tabular}{lc}
\toprule
\textbf{Statistic} & \textbf{Value} \\
\midrule
Total \#Tasks & 15{,}096 \\
Avg. \#Tools Per Task & 4.1 \\
Avg. \#Interactions (After Training) & 13.4 \\
Avg. Mapping Size $|M|$ (After Training) & 30.1 \\
Avg. Per‑Task Token Usage in Synthesis & 25{,}278 \\
\bottomrule
\end{tabular}
}
\caption{Statistics of the synthesized tool-use dataset.}
\label{tab:synthagent-data-stats-vertical}
\end{table}

\begin{table*}[t]
\centering
\setlength{\tabcolsep}{3pt}
\resizebox{0.975\textwidth}{!}{
\begin{tabular}{ccccccccc}
\toprule
\multirow{2}{*}{Method} &
\multicolumn{3}{c}{TAU-2 Bench} &
\multicolumn{4}{c}{BFCL-V4-Multi-turn} &
\multirow{2}{*}{Avg.} \\
\cmidrule(lr){2-4} \cmidrule(lr){5-8}
& Airline & Telecom & Retail &
Base & Miss Func & Miss Param & Long Context & \\
\midrule

\multicolumn{9}{c}{\fontsize{10}{9.5}\selectfont\textbf{Baselines trained on closed-source or the latest open-source data (non-thinking, using tools)}} \\
% \multicolumn{9}{c}{(non-thinking, using tools)}
\midrule
Qwen3-235B          & \textcolor{gray}{47.5} & 37.7 & \textcolor{gray}{68.0} & \textcolor{gray}{58.5} & \textcolor{gray}{47.5} & \textcolor{gray}{35.0} & \textcolor{gray}{54.0} & \textcolor{gray}{49.7} \\
Qwen3-14B           & 22.0 & 25.4 & 39.5 & 40.0 & 34.5 & 26.5 & 26.5 & 30.6 \\
Qwen3-32B           & 22.5 & 27.6 & 44.7 & 50.5 & 43.0 & 30.5 & 33.0 & 36.0 \\
ToolStar-8B         & 13.5 & 25.9 & 39.5 & 52.0 & 38.0 & 22.5 & 30.5 & 31.7 \\
ToolStar-14B        & 18.0 & 30.7 & 40.4 & 56.5 & 35.5 & 29.5 & 39.5 & 35.7 \\
% \midrule
% \multicolumn{9}{c}{\textbf{RL-trained (non-thinking, using tools)}} \\
\midrule

\rowcolor{blue!10}  \textsc{SynthAgent}-8B  & 34.5 & 38.2 & 57.2 & 54.5 & 45.5 & \textbf{33.0} & 37.5 & 42.9 \\
\rowcolor{blue!10}  \textsc{SynthAgent}-14B & \textbf{40.0} & \textbf{44.7} & \textbf{58.6} & \textbf{57.0} & \textbf{46.5}& 31.0 & \textbf{46.0} & \textbf{46.3} \\

\bottomrule
\end{tabular}
}
\caption{Agentic performance comparison. 
For TAU-2 (Airline, Telecom, Retail), we report Avg@4 and use the open-source Kimi-K2-Instruct model as the user simulator. Qwen3-235B refer to the model Qwen3-235B-A22B-Instruct. The same applies in the following text. Best results except Qwen3-235B are \textbf{bolded}.
% the limited number of examples makes Avg@4 a more stable metric, so
% , which may slightly differ from the public leaderboard results.
}
\label{tab:tau2_bfcl}
\end{table*}

\begin{table*}[t]
\centering
\setlength{\tabcolsep}{3pt}
\resizebox{0.975\textwidth}{!}{
\begin{tabular}{ccccccccc}
\toprule
\multirow{2}{*}{Method} &
\multicolumn{4}{c}{Math} &
\multicolumn{3}{c}{Search} &
\multirow{2}{*}{Avg.} \\
\cmidrule(lr){2-5} \cmidrule(lr){6-8}
& AIME24 & AIME25 & Olympiad & HMMT25
& Frames & XBench & WebWalker & \\
\midrule

% \multicolumn{9}{c}{\textbf{Qwen3 Baselines (non-thinking, using tools)}} \\
\multicolumn{9}{c}{\fontsize{10}{9.5}\selectfont\textbf{Baselines trained on closed-source or the latest open-source data (non-thinking, using tools)}} \\
\midrule
Qwen3-235B              & \textcolor{gray}{83.3} & \textcolor{gray}{70.6} & \textcolor{gray}{83.4} & \textcolor{gray}{64.4} & \textcolor{gray}{70.5} & 43.0 & \textcolor{gray}{59.5} & \textcolor{gray}{67.8} \\
Qwen3-14B               & 39.4 & 35.6 & 53.8 & 39.4 & 37.8 & 21.0 & 38.0 & 37.9 \\
Qwen3-32B               & 50.0 & 41.1 & 56.4 & 35.1 & 44.8 & 25.0 & 37.0 & 41.3 \\
ToolStar-8B             & 60.6 & 54.4 & 76.4 & 47.8 & 58.5 & 33.0 & 44.5 & 53.6 \\
ToolStar-14B            & 71.7 & 63.3 & 77.3 & 45.0 & 60.4 & 40.0 & 44.0 & 57.4 \\

\midrule
% \multicolumn{9}{c}{\textbf{RL-trained (non-thinking, using tools)}} \\

\rowcolor{blue!10} \textsc{SynthAgent}-8B  & 71.6 & 58.9 & 77.2 & 48.9 & 59.7 & \textbf{45.0} & 48.5 & 58.5 \\
% \rowcolor{blue!10} \textsc{SynthAgent}-14B & 72.2 & 66.7 & 80.1 & 53.9 & 50.0 & 34.0 & 42.0 & 57.0 \\
\rowcolor{blue!10} \textsc{SynthAgent}-14B & \textbf{72.2} & \textbf{66.7}& \textbf{80.1} & \textbf{53.9} & \textbf{63.5} & 43.0 & \textbf{50.0} & \textbf{61.3} \\

\bottomrule
\end{tabular}
}
\caption{
Short-horizon reasoning performance comparison (math and search). 
For AIME24, AIME25, and HMMT25, we report Avg@6 for more stable evaluation. Best results except Qwen3-235B are \textbf{bolded}.
}
\label{tab:reasoning_deepsearch}
\end{table*}

% \begin{table*}[t]
% \centering
% \setlength{\tabcolsep}{3pt}
% \resizebox{0.975\textwidth}{!}{
% \begin{tabular}{ccccccccc}
% \toprule
% \multirow{2}{*}{Method} &
% \multicolumn{3}{c}{TAU-2 Bench} &
% \multicolumn{4}{c}{BFCL-Multi-turn} &
% \multirow{2}{*}{Avg.} \\
% \cmidrule(lr){2-4} \cmidrule(lr){5-8}
% & Airline & Telecom & Retail &
% Base & Miss Func & Miss Param & Long Context
% \\
% \midrule

% Qwen3-8B &
% 15.0 & 8.8 & 32.5 &
% 35.5 & 35.0 & 20.5 & 21.5 &
% 24.1
% \\

% 8B, reasoning data only &
% 22.0 & 26.1 & 38.2 &
% 52.5 & 43.5 & 25.5 & 30.0 &
% 34.0
% \\

% 8B, tool-use data &
% 29.5 & 32.7 & 57.2 &
% 54.5 & 45.5 & 33.0 & 37.5 &
% 41.4
% \\

% \bottomrule
% \end{tabular}
% }
% \caption{Ablation study on agentic benchmark. We report results across all 7 agentic tool-use datasets. 
% % We synthesized a large set of diverse tool-use data along with a smaller amount of math-reasoning data. This ablation examines how different components of the synthetic dataset contribute to the model's agentic performance.
% }
% \label{tab:ablation_tau2_bfcl}
% \end{table*}

\subsection{Experimental Setup}
We evaluate \textsc{SynthAgent} by assessing models trained on our synthetic data and within simulated environments. Our experiments focus on agentic benchmarks that measure long-horizon tool use, multi-turn planning, and adaptability to unfamiliar tools. We also test short-horizon generalization through reasoning tasks such as math and search.

\paragraph{Agentic Tool Use Benchmarks}

We evaluate on the most representative agentic benchmarks: TAU-2~\cite{barres2025tau2} and BFCL-V4~\cite{patil2025bfcl}, spanning 7 datasets and nearly 100 diverse tools. These benchmarks are widely used by Qwen~\cite{yang2025qwen3}, Kimi~\cite{team2025kimi}, and DeepSeek~\cite{deepseekai2025deepseekv32pushingfrontieropen}, aligning our protocol with that of leading foundation models.

BFCL-V4 provides multiple datasets; we focus on its multi-turn subset (about 800 tasks). These tasks span diverse real-world domains such as trading, vehicle control, and social media. Each task typically requires 5 to 20 tool-interaction turns, providing a rigorous evaluation of the model's capabilities in parameter clarification and error rejection.

TAU-2 targets three real-world business domains: airline, retail, and telecommunications, comprising roughly 300 tasks. These tasks generally require multi-turn interactions between the agent and user. Moreover, users can also invoke tools and modify the environment, meaning the model must not only execute tools correctly but also guide the user and handle uncertain feedback.

The above agentic benchmarks, with their unfamiliar tools and long-horizon planning demands, serve as our primary evaluation suite.

\paragraph{Reasoning Benchmarks}
We also examine the short-horizon reasoning capabilities of our framework. We employ several math benchmarks (AIME24, AIME25, HMMT25, Olympiad~\cite{he-etal-2024-olympiadbench}) and search benchmarks (FRAMES~\cite{krishna-etal-2025-fact}, WebWalker~\cite{wu2025webwalker}, XBench~\cite{chen2025xbench}). These tasks involve only two tools: a Python interpreter and Google Search and typically require fewer than five interaction turns. However, each step demands deeper reasoning than in the agentic benchmarks, making them suitable for out-of-domain evaluation.

All benchmarks were introduced after 2024, ensuring strong relevance and up-to-date evaluation.

\paragraph{Evaluation}
TAU-2 and BFCL provide not only datasets but also full interactive environments. During evaluation, the model must invoke tools to interact with these environments; performance is measured by checking whether environment states are correctly updated to their ground-truth values using Exact Match. For math reasoning, we also apply Exact Match. For more free-form outputs in search reasoning, we use Qwen3-235B to judge whether the model's responses are semantically consistent with the ground truth. The search tool is implemented via the Google Search API.

\paragraph{Baselines}
We compare our model, trained on synthetic tasks and mock environments, against the following baselines: RL-trained models on the latest open-source ToolStar~\cite{dong2025toolstarempoweringllmbrainedmultitool}, which employ 30,000 math and search examples. We also evaluate strong LLMs prompted to integrate tools, such as the larger Qwen3-32B. The Qwen3 technical report~\cite{yang2025qwen3} indicates that Qwen3-32B has already been trained on synthetic tool-use data using RL, making it a competitive baseline. All baselines perform inference using the OpenAI function-calling format and the same prompt, ensuring a fully consistent setup.

% Since open-source synthetic tool-use datasets are scarce, we use the latest ToolStar~\cite{dong2025toolstarempoweringllmbrainedmultitool}, which contains about 60,000 math and search examples, and sample 30,000 instances from it for RL training. We also evaluate strong LLMs prompted for tool integration, such as the larger Qwen3-32B. Qwen3 technical report indicates that it has already been trained on synthetic tool-use data, making it a competitive baseline.

% evaluating whether capabilities learned solely from synthetic tasks and simulated environments can generalize to real agentic benchmarks.

\paragraph{Implementation}
% 4,000 math or search reasoning tasks 
Using \textsc{SynthAgent}, we generate 15,096 synthetic tool-use tasks entirely with locally deployed open-source LLMs. We train Qwen3-8B/14B with GRPO (non-thinking) to assess data quality. For rubric construction, we collect four demonstrations from a strong agentic teacher (Qwen3-235B). Since rubric design depends mainly on high-level workflow rather than specific teacher demonstrations, the number of them typically has little impact, as shown in Section~\ref{sec:futher}. For reward, we employ Qwen3-30B-A3B-
Instruct to judge with rubrics. More implementation details are provided in Appendix~\ref{sec:implementation}.

% Using \textsc{SynthAgent}, we built a fully synthetic training dataset with Qwen3-235B and carefully designed prompts. In total, we generated 15,096 tool-use tasks, along with a small number of synthetic math reasoning tasks (around 4,000). All data were produced locally with open-source LLMs, without relying on commercial APIs, ensuring a stable and cost-efficient pipeline. To assess data quality, we trained Qwen3-8B and Qwen3-14B using GRPO and adopted the non-thinking mode to reduce training time.

% For teacher demonstrations, we employed two strong agentic teacher models: Qwen3-235B-A22B and Kimi-K2-Instruct. Each teacher repeatedly solved the same synthetic tasks twice, producing diverse trajectories essential for constructing reliable task-level rubrics. Additional dataset descriptions, prompt templates, and implementation details are provided in Appendix~\ref{sec:implementation}.

\begin{figure*}[t]
\centering
\begin{minipage}[t]{0.45\textwidth}
    \centering
    \includegraphics[width=\textwidth]{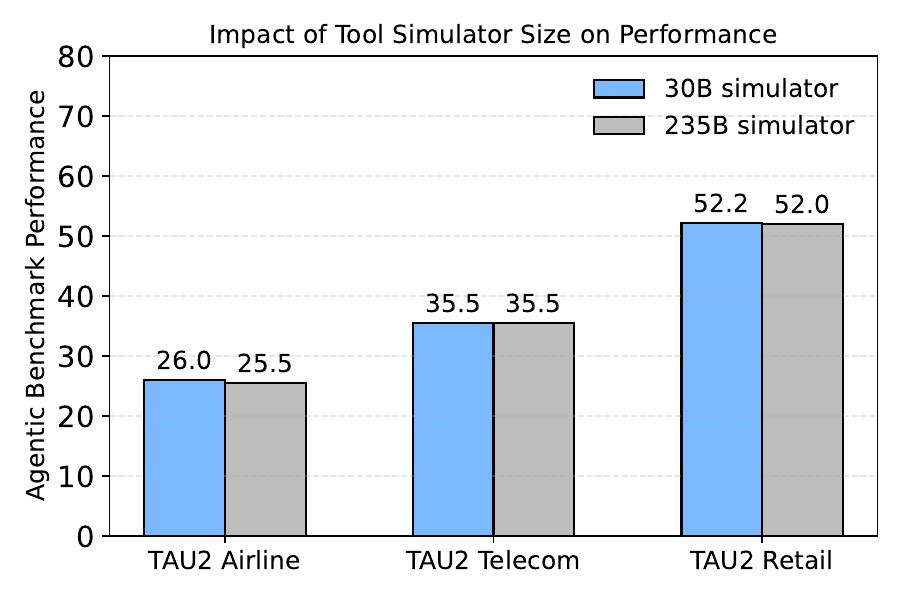}
    \caption{Effect of tool-simulator size on TAU-2 performance(5,000 training samples), showing negligible gains from larger simulators.}
    \label{fig:simulator_comparison}
\end{minipage}
\hfill
\begin{minipage}[t]{0.52\textwidth}
    \centering
    \includegraphics[width=\textwidth]{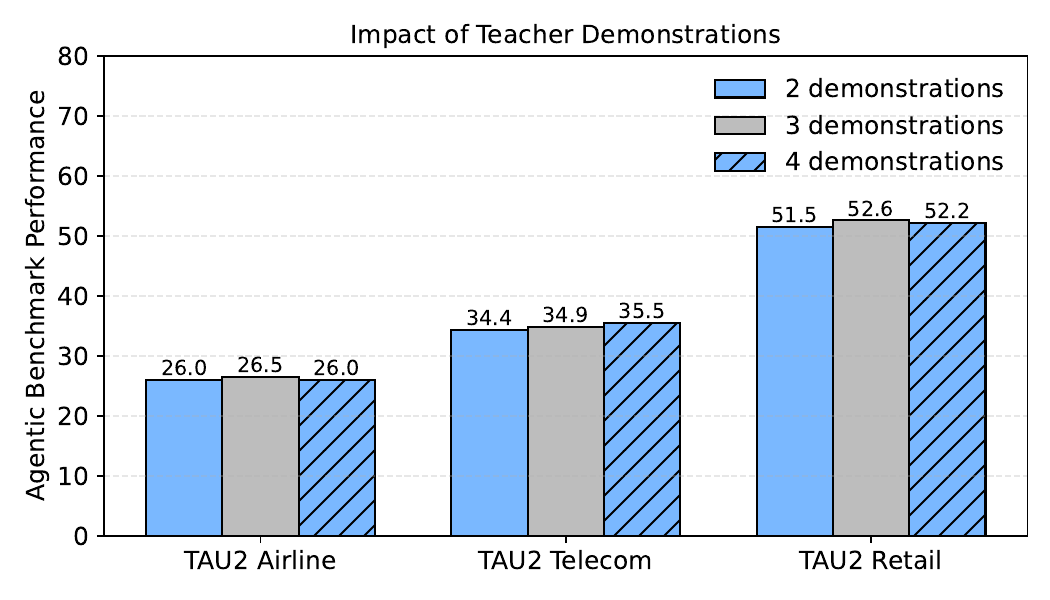}
    \caption{Influence of teacher-demonstration count when constructing task-level rubrics(5,000 training samples), indicating that additional demonstrations yield limited gain.}
    \label{fig:teacher_demonstrations}
\end{minipage}
\end{figure*}

\subsection{Main Results}
Table~\ref{tab:tau2_bfcl} reports the performance of our 8B and 14B models, trained on synthetic data and simulated environments, on real-world agentic benchmarks such as TAU-2 and BFCL-Multi-turn.

\noindent\textbf{\textsc{SynthAgent} enables small models to match and even surpass much larger agentic models.}  
Using synthetic tool-use tasks and fully simulated environments, our method yields substantial gains across real-world agentic benchmarks. On TAU-2 and BFCL-Multi-turn, \textsc{SynthAgent}-8B scores 42.9 on average (+12.3 over Qwen3-14B, and +6.9 over Qwen3-32B). The improvements are even greater with \textsc{SynthAgent}-14B, which scores 46.3 on average.
% (+14.6 over Qwen3-14B, +10.7 over Qwen3-32B). 
Despite being far smaller, \textsc{SynthAgent}‑14B matches Qwen3‑235B on many TAU‑2 and BFCL domains, demonstrating that our synthetic training strategy can close the gap with much larger models.

\noindent\textbf{Synthetic data and simulated environments substantially outperform existing open-source datasets.}  
Open-source agentic datasets cover only a limited set of tools and cannot capture the complexity of real tool-use logic. For example, in the latest ToolStar~\cite{dong2025toolstarempoweringllmbrainedmultitool} dataset, RL-trained models show no significant improvement in agentic performance. In contrast, \textsc{SynthAgent}-8B achieves great performance in TAU-2, consistently surpassing open-source baselines, with BFCL-Multi-turn showing the same pattern. These results indicate that diverse and challenging synthetic tasks are far more effective for strengthening a model's agentic abilities.

\noindent\textbf{Beyond tool-use tasks, the model also generalizes to short-horizon reasoning.}  
Table~\ref{tab:reasoning_deepsearch} shows that, although math and search tasks are not the primary focus of our training(only 4,000 instances sampled from ToolStar are included), \textsc{SynthAgent} still achieves substantial gains. Under the same non‑thinking setting, \textsc{SynthAgent}-8B significantly outperforms Qwen3‑14B and even exceeds an 8B model trained on 30,000 ToolStar examples. The improvements for \textsc{SynthAgent}-14B are even larger. These results indicate that our method transfers effectively to new reasoning domains, with tool-use data also exhibiting strong generalization in reasoning tasks.

Overall, training on synthetic tasks and environments allows 8B–14B models to rival or surpass 32B models, drastically reducing inference cost.

\begin{table}[t]
\centering
\setlength{\tabcolsep}{1pt}
\resizebox{0.5\textwidth}{!}{
\begin{tabular}{cccccc}
\toprule
\multirow{2}{*}{Method} &
\multicolumn{3}{c}{TAU-2 Bench} &
\multirow{2}{*}{\shortstack{BFCL-V4\\Multi-turn\\Avg.}} \\
\cmidrule(lr){2-4}
& Airline & Telecom & Retail \\
\midrule

Qwen3-8B & 15.0 & 8.8 & 32.5 & 28.1 \\

reasoning data only & 22.0 & 26.1 & 38.2 & 37.9 \\

W/O information gap & 22.0 & 28.9 & 35.3 & 39.1 \\
\midrule
\textsc{SynthAgent}-8B & 34.5 & 38.2 & 57.2 & 42.6 \\
\bottomrule
\end{tabular}
}
\caption{Ablation study on agentic benchmarks.}
\label{tab:ablation_tau2_bfcl}
\end{table}

\subsection{Further Analysis}
\label{sec:futher}
% The effectiveness of agentic RL depends on high-quality, diverse data. Ablation in Table~\ref{tab:ablation_tau2_bfcl} confirms that synthetic tool-use data is essential for agentic performance. We compare three settings: the Qwen3-8B baseline, training on synthetic reasoning-only data, and training on our full synthetic tool-use data. Reasoning-only data provides a modest improvement over the Qwen3-8B baseline, mainly enhancing short-horizon reasoning. However, it remains insufficient for multi-turn repair, parameter recovery, or robust tool grounding. In contrast, synthetic tool-use data yields a much larger boost, substantially improving performance across all agentic benchmarks.

\noindent\textbf{Ablation} The results in Table~\ref{tab:ablation_tau2_bfcl} confirm that synthetic tool-use data is crucial. Using only reasoning data yields improvements over the Qwen3-8B baseline, mainly from short-term reasoning, but it remains inadequate for multi-turn long-horizon agentic tasks. In contrast, adding synthetic tool-use data provides substantial gains and consistently improves performance across all agentic benchmarks. However, if we do not introduce information gaps, do not rewrite workflows into less explicit descriptions, and directly use them for training without user interaction, the benefit of tool-use data becomes negligible. This further validates the rationale behind our design.

\noindent\textbf{Impact of Tool Simulator Size on Performance} The tool simulator primarily generates responses to new tool calls, and checks whether a call matches a previous query in the prompt. Both are simple, well-defined operations that typically do not require a strong model. To validate this, we evaluate Qwen3-8B trained on 5,000 synthetic agentic tasks with two simulators: Qwen3-235B and the smaller Qwen3-30B-A3B-Instruct model. As shown in Figure~\ref{fig:simulator_comparison}, a larger simulator does not improve performance, suggesting mock tool simulation is largely formatted QA and semantic matching, rather than a capability that benefits from model scale.

\noindent\textbf{Impact of Number of Teacher Demonstrations} The rubrics and subgoals are mainly derived from the high-level workflow in Section~\ref{sec:tool_use_task_generation}. Teacher trajectories are used only to align the synthesized workflow with real executions; examples where the teacher covers too little of the workflow are discarded. In principle, rubric quality depends weakly on the number of teacher trajectories.

To evaluate this, we build rubrics for 5,000 synthetic agentic examples using 2, 3, or 4 teacher trajectories and compare performance. As shown in Figure~\ref{fig:teacher_demonstrations}, adding more demonstrations yields no significant gains, suggesting rubric construction does not require substantial computation.

More experiments, such as data-scaling effects on RL and RL-SFT comparisons at equal data sizes, are provided in Appendix~\ref{sec:additional_experiments}.

\section{Conclusion}

We present \textsc{SynthAgent}, a novel framework addressing two core bottlenecks in training agentic language models: the scarcity of diverse, challenging tasks and stable tool environments. By jointly synthesizing tool-use tasks with underspecified instructions and providing stable mock environments, \textsc{SynthAgent} enables efficient reinforcement learning for small models. Extensive evaluations demonstrate that models trained entirely on synthetic data and virtual environments achieve substantial gains, with small models surpassing much larger baselines.

\section*{Limitations}

% While \textsc{SynthAgent} provides a cost-effective framework for training agentic language models, several limitations remain. First, although synthetic tasks and simulated environments facilitate diverse data generation and stable training, they may not fully capture the complexity and unpredictability of real-world interactions, potentially limiting generalization in highly dynamic or novel scenarios. Second, the quality of the generated tasks and tool simulations depends heavily on the capabilities of the teacher LLMs; any biases or shortcomings in these models may propagate to the synthetic data, consequently affecting downstream agent performance. Finally, although our reward design based on teacher trajectories offers objective signals, it may not fully encompass all desirable aspects of task success, such as nuanced contextual understanding. Future work should explore extending to richer environment dynamics and incorporating more sophisticated evaluation criteria to address these challenges.
Agentic training data synthesis is an increasingly important research topic. Technical reports from leading foundation models (Qwen3~\cite{yang2025qwen3}, LongCat~\cite{team2025longcat}, Kimi K2~\cite{team2025kimi}, DeepSeek V3.2~\cite{deepseekai2025deepseekv32pushingfrontieropen}, Minimax M2~\cite{chen2025minimax}) consistently show that synthetic data, rather than real-world data, forms the core of agentic RL. However, these models do not release their synthetic datasets, nor do they provide detailed descriptions of their synthesis procedures. This limitation restricts our ability to refine the \textsc{SynthAgent} pipeline based on prior work and makes it difficult to compare against stronger baselines. Future work should explore additional approaches for agentic training data synthesis and identify the key factors that are most critical for building effective agents.

\bibliography{anthology,custom}

\appendix
\section*{The Use of Large Language Models~(LLMs) in Writing}

An LLM (specifically OpenAI's GPT‑5~\cite{achiam2023gpt}) was used solely for minor language editing, including grammar correction and light rephrasing for clarity. It did not contribute to the research design, and all scientific content is entirely the authors' own.

\begin{figure*}[t]
\centering
\begin{minipage}[t]{0.60\textwidth}
    \centering
    \includegraphics[width=\textwidth]{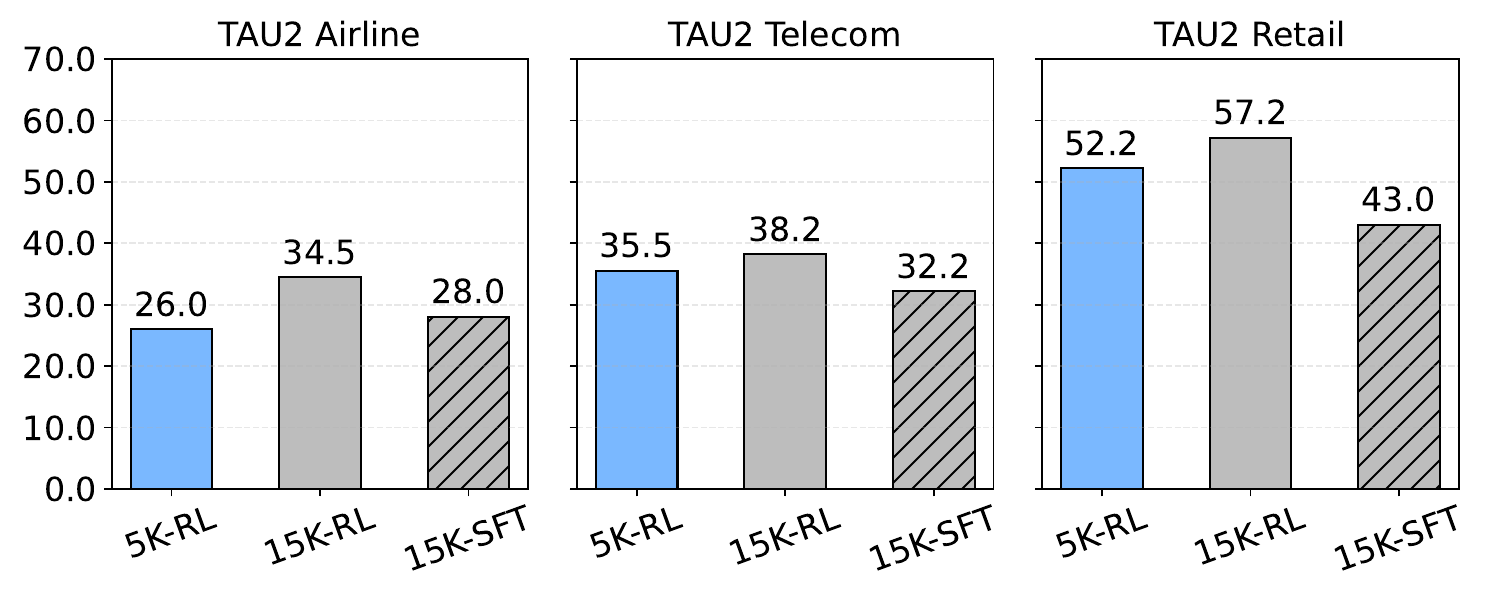}
    \caption{Impact of increased training data on RL performance, and comparison between RL and SFT at the same data scale.}
    \label{fig:scale}
\end{minipage}
\hfill
\begin{minipage}[t]{0.39\textwidth}
    \centering
    \includegraphics[width=\textwidth]{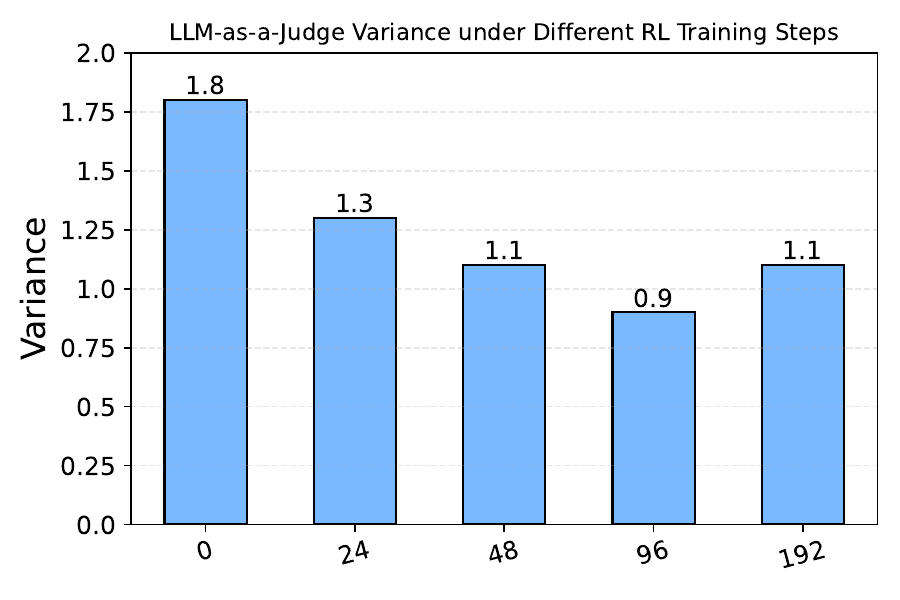}
    \caption{Stability analysis of LLM-as-a-Judge: score variance on the sample tasks across different training steps.}
    \label{fig:variance}
\end{minipage}
\end{figure*}

\section{More Implementation Details}
\label{sec:implementation}

\paragraph{Synthetic Data}  
All personas used during data synthesis are exclusively sourced from Persona Hub~\cite{ge2025scalingsyntheticdatacreation}. During toolset generation, we define virtual tools following the OpenAI function-calling specification and filter out any samples that cannot be parsed into a valid function-calling format. In total, we generated 15,096 tool-use tasks, along with a smaller set of math or search reasoning tasks (approximately 4,000). All data were produced locally with open-source LLMs, without reliance on commercial APIs, ensuring a stable and cost-efficient pipeline.

\paragraph{Benchmarks}  
Detailed benchmark statistics (e.g., subsets and test sizes) are reported in Table~\ref{tab:benchmarks}.  
Agentic tool-use benchmarks (TAU-2 and BFCL-V4 Multi-turn) measure long-horizon, multi-turn interaction with unfamiliar tools. Reasoning benchmarks (math and search) evaluate out-of-domain generalization, where tasks typically involve fewer tool calls but require deeper multi-step reasoning and evidence synthesis.

\paragraph{Training}  
We perform reinforcement learning using the GRPO algorithm~\citep{shao2024deepseekmath} within the VERL framework~\citep{sheng2024hybridflow}. We use a global batch size of 128, a PPO mini-batch size of 16, a rollout size of 16, and a maximum response length of 13,000 tokens, training for 2 epochs on 8$\times$NVIDIA H20 GPUs. The clipping ratio is constrained between 0.2 and 0.28, and we allow up to 16 turns per rollout. Since Qwen3 base models already exhibit strong tool/function-calling capabilities and reliably follow the required format, we skip the SFT phase for format learning and directly proceed with RL training.

\paragraph{Inference}  
During inference, the model was deployed with SGLang~\citep{zheng2024sglang} to increase throughput. TAU-2~\citep{barres2025tau2} and BFCL~\citep{patil2025bfcl} provide complete evaluation code, and we follow their official settings. All baseline settings, including temperature, max steps, system prompts, sampling strategies, and tool formats, match the official evaluation code exactly. For factual reasoning datasets, the search tool is implemented using the Google Search API, with country set to ``us'' and top-$k$ set to 5. We use only the text snippets returned by the API as observations, omitting full web browser outputs or long context summaries.

BFCL-V4 and TAU2 are continuously evolving agentic benchmarks, which replace older test data with more challenging samples. This makes previously reported results potentially outdated. For consistency, we re-evaluate all baselines using the latest BFCL-V4 and TAU2 versions available as of December 2025.

\begin{table*}[t]
\centering
\renewcommand{\arraystretch}{1.2}
\small
\setlength{\tabcolsep}{5pt}
\begin{threeparttable}
\begin{tabular}{m{2.5cm} m{3.5cm} m{2.0cm} p{5.3cm}}
\toprule
\textbf{Category} & \textbf{Benchmark / Subset} & \textbf{Test Size} & \textbf{Description} \\
\midrule

\multirow{15}{*}{\centering Agentic Tool Use}
& TAU-2: Airline~\citep{barres2025tau2} (Avg@4) & 50 & Airline booking and service workflows; multi-turn tool use with user interventions. \\
& TAU-2: Telecom (Avg@4) & 114 & Telecom troubleshooting and account operations; user can also call tools. \\
& TAU-2: Retail (Avg@4) & 114 & Retail returns and order management; multi-step tool execution in dialogs. \\

\cmidrule(lr){2-4}

& BFCL-V4: Multi-turn / Base~\citep{patil2025bfcl} & 200 & Multi-turn function calling across domains; end-to-end tool orchestration. \\
& BFCL-V4: Multi-turn / Miss Func & 200 & Missing or invalid functions; tests tool rejection and plan adjustment. \\
& BFCL-V4: Multi-turn / Miss Param & 200 & Missing required arguments; tests parameter elicitation and correction. \\
& BFCL-V4: Multi-turn / Long Context & 200 & Long-context dialogs; tests memory and consistency over many turns. \\

\midrule
\multirow{8}{*}{\centering Math Reasoning}
& AIME24$^{1}$  (Avg@6) & 30 & 2024 AIME math problems in algebra and geometry; assesses advanced reasoning. \\
& AIME25$^{2}$  (Avg@6) & 30 & 2025 AIME I\&II across major topics; evaluates out-of-domain math reasoning. \\
& OlympiadBench~\citep{he-etal-2024-olympiadbench} & 674 & Olympiad-level math problems; tests hard multi-step reasoning. \\
& HMMT25$^{3}$  (Avg@6) & 30 & Recent contest math problems; evaluates robustness on new distributions. \\

\midrule
\multirow{6}{*}{\centering Search Reasoning}
& FRAMES~\citep{krishna-etal-2025-fact} & 824 & Search-based QA with evidence synthesis; evaluates factual reasoning under retrieval. \\
& xBench~\citep{chen2025xbench} & 100 & Deep-search benchmark; multi-hop exploration and cross-source synthesis. \\
& WebWalker~\citep{wu2025webwalker} & 200 & Web navigation and retrieval; multi-step searching in dynamic settings. \\
\bottomrule
\end{tabular}
\begin{tablenotes}
\item[1] \url{https://huggingface.co/datasets/HuggingFaceH4/aime_2024}
\item[2] \url{https://huggingface.co/datasets/math-ai/aime25}
\item[3] \url{https://huggingface.co/datasets/MathArena/hmmt_feb_2025}
\end{tablenotes}
\end{threeparttable}
\caption{Overview of evaluation benchmarks.}
\label{tab:benchmarks}
\end{table*}

\section{Additional Experiment Results}
\label{sec:additional_experiments}
Figure~\ref{fig:scale} illustrates how increasing training data affects agentic performance on the TAU-2 benchmark. Models trained with 15K tool-use samples clearly outperform those trained with 5K, highlighting the quality of our synthetic data and suggesting that scaling synthetic data and RL compute can further enhance agentic capabilities. The figure also compares RL and SFT under the same data scale. The SFT trajectories are synthesized by Qwen3-235B; however, SFT yields much smaller gains than RL, as RL can generate far more diverse trajectories through exploration, which is a crucial factor for improving agentic behavior.

During RL training, we use an LLM to assign rewards to trajectories based on our synthesized rubrics. These rubrics guide the evaluation process, substantially reducing LLM-as-a-Judge variance. We sample 100 tasks and generate trajectories for each task at different RL training steps, scoring them with Qwen30B-A3B-Instruct and computing the mean. Figure~\ref{fig:variance} reports the percentage score variance (repeat 6 times) across training steps. From the untrained model to near convergence, the LLM's scores on the same tasks become increasingly consistent, demonstrating the stability of our rubric-based RL training.

\section{Synthetic Task and Tool Examples}
\label{sec:example}
In Table~\ref{tab:tooluse_case}, we provide examples of the synthetic data, including task descriptions, user-only information, and the tool set formatted according to the OpenAI function-calling schema.

Table~\ref{tab:rubric_case} presents the rubrics used to evaluate task completion, covering constraints, sub-goals, and interactions between the user and the agent.

Table~\ref{tab:interaction_case} illustrates the complete execution flow of our model operating in simulated environments, where it invokes tools, interacts with the user, and solves problems throughout the process.

To ensure full reproducibility, we release the entire data-generation code, including all prompts used for synthesis, the synthetic dataset, and the reinforcement learning code based on this data. All resources are available at the following link: \url{https://anonymous.4open.science/r/SYNTHAGENT-68A4/}. 

All prompts used in the paper can be found in the anonymous repository under the path \texttt{tool\_use\_data\_synthesis/functions}.

\begin{table*}[ht]
\centering
\renewcommand{\arraystretch}{1.12}
\begin{tabular}{p{0.97\textwidth}}
\toprule
% \rowcolor{orange!20}
\textbf{Fuzzy Task:} 
Create a weekly digital newsletter for a senior resident that shares personal updates, meaningful stories, and family prompts in an accessible and heartfelt format. \\
\midrule
% \rowcolor{orange!20}
\textbf{User Only:} 
The senior prefers speaking over typing and can comfortably record short voice messages of up to 90 seconds. The newsletter should reflect their voice and experiences, include a photo from their life, and feature a thoughtful quote related to family or memory. Each edition must end with a gentle question to invite responses from younger family members. The final version must be delivered as a well-formatted email that displays correctly on phones and tablets, with optional audio support for low-vision recipients. \\
\midrule

% \rowcolor{orange!20}
\textbf{Constraint:} 
The agent must not use \texttt{VoiceTranscriber} on audio longer than 90 seconds. A violation occurs if the agent transcribes without confirming duration, assumes the audio is short enough, or proceeds when the audio is known/implied to exceed 90 seconds. \\

\midrule 

% \rowcolor{orange!20} 
\textbf{Tools (OpenAI function-calling spec):} \\

\textbf{VoiceTranscriber} \\
\textit{Description:} Converts short spoken audio clips into punctuated text; designed for seniors with a simple UI. \\
\textit{Parameters:} \\
\hspace*{1em}\texttt{audio\_file\_path} (string): File path or URL to the audio clip (\texttt{.mp3}/\texttt{.wav}). \\

\textbf{LiteratureFinder} \\
\textit{Description:} Retrieves a short quote or poem excerpt from a curated public-domain database given a theme keyword. \\
\textit{Parameters:} \\
\hspace*{1em}\texttt{search\_query} (string): Keyword/theme (e.g., \texttt{family}, \texttt{memory}, \texttt{grandchild}). \\

\textbf{NewsletterFormatter} \\
\textit{Description:} Assembles text, image, and quote into an accessible HTML email template with responsive layout and an optional audio link. \\
\textit{Parameters:} \\
\hspace*{1em}\texttt{headline} (string): Main title/subject line of the newsletter. \\
\hspace*{1em}\texttt{body\_text} (string): Main content paragraph (e.g., weekly update). \\
\hspace*{1em}\texttt{image\_url} (string): URL to the featured image. \\
\hspace*{1em}\texttt{quote} (string): Curated quote or poem excerpt to include. \\
\hspace*{1em}\texttt{attribution} (string): Attribution formatted as \texttt{``Author, Title''}. \\
\hspace*{1em}\texttt{question\_of\_week} (string): Prompt question to encourage family responses. \\
\hspace*{1em}\texttt{audio\_narration\_url} (string, optional): URL to an audio version of the newsletter. \\

\textbf{ImageDescriber} \\
\textit{Description:} Generates accessibility-oriented alt-text for a given image URL. \\
\textit{Parameters:} \\
\hspace*{1em}\texttt{image\_url} (string): URL of the image to be described. \\

\textbf{AudioNarrationGenerator} \\
\textit{Description:} Simulates converting written text into an audio narration file for accessibility. \\
\textit{Parameters:} \\
\hspace*{1em}\texttt{text} (string): The text content to be narrated. \\
\bottomrule
\end{tabular}
\caption{An example of our synthesized tool-use training data, including the underspecified instruction (\texttt{fuzzy task}) and the OpenAI function-calling style tool specifications (\texttt{tools}).}
\label{tab:tooluse_case}
\end{table*}

\begin{table*}[t]
\centering
\renewcommand{\arraystretch}{1.05}
\begin{tabular}{p{0.98\textwidth}}
\toprule
\textbf{Rubric} \\
\midrule

\textbf{1. Forbidden Behaviors} \\
\textbf{Duration constraint:} The agent must not use \texttt{VoiceTranscriber} on audio longer than 90 seconds. A violation occurs if the agent transcribes without confirming duration, assumes the audio is short enough, or proceeds when the audio is known/implied to exceed 90 seconds. \\

\midrule
\textbf{2. Task Sub-goals} \\
(1) Obtain a $\le$90s voice update and confirm duration. \\
(2) Transcribe the voice message via \texttt{VoiceTranscriber}. \\
(3) Retrieve a relevant quote via \texttt{LiteratureFinder}. \\
(4) Include a user-provided image and generate alt-text via \texttt{ImageDescriber}. \\
(5) Add a family prompt question. \\
(6) Generate audio narration via \texttt{AudioNarrationGenerator}. \\
(7) Format the newsletter via \texttt{NewsletterFormatter}. \\

\midrule
\textbf{3. Required User Interactions (must ask/confirm before acting)} \\
(1) Voice message availability, file path, and whether it is under 90 seconds. \\
(2) Image availability and a valid image URL. \\
(3) Newsletter theme/focus (memory/story to highlight). \\
(4) Audience and tone preferences. \\
(5) Whether audio narration is desired/needed. \\
\bottomrule
\end{tabular}
\caption{An example rubric for the synthesized training data instance in Table~\ref{tab:tooluse_case}, including forbidden behaviors, task sub-goals, and required user interactions.}
\label{tab:rubric_case}
\end{table*}

\begin{table*}[t]
\centering
\small
\renewcommand{\arraystretch}{1.18}
\setlength{\tabcolsep}{6pt}
\begin{tabular}{p{0.09\textwidth} p{0.86\textwidth}}
\toprule
\textbf{Role} & \textbf{Content} \\
\midrule

\rowcolor{gray!10}
User & Create a weekly digital newsletter for a senior resident that shares personal updates, meaningful stories, and family prompts in an accessible and heartfelt format. (Ask for clarification if required parameters are missing.) \\

Assistant & Requests missing details (personal update, theme, image URL, audio narration, target audience). \\

\rowcolor{gray!10}
User & The senior recorded a short voice message about planting flowers in the garden with a grandchild---would you like me to transcribe that as the personal update? \\

Assistant & Confirms using the voice message and asks for two required details: (i) audio length $\le$ 90 seconds, and (ii) an \texttt{.mp3} or \texttt{.wav} file link/path. \\

\rowcolor{gray!10}
User & Yes, it is 85 seconds and here is the \texttt{.mp3} link: \texttt{https://example.com/audio/flower\_garden.mp3}. \\

\rowcolor{blue!6}
Assistant (Tool Call) &
\texttt{VoiceTranscriber(audio\_file\_path="https://example.com/audio/flower\_ garden.mp3")} \\

\rowcolor{green!6}
Tool (Response) &
\texttt{\{"transcribed\_text": "Every spring, I would plant marigolds along the garden path... I feel her joy all over again.", "confidence\_score": 0.96\}} \\

Assistant & Plans to add a complementary literary quote and calls \texttt{LiteratureFinder} with a relevant theme. \\

\rowcolor{blue!6}
Assistant (Tool Call) &
\texttt{LiteratureFinder(search\_query="garden and memory")} \\

\rowcolor{green!6}
Tool (Response) &
\texttt{\{"quote": "They flash upon that inward eye ...", "attribution": "--- William Wordsworth, from "I Wandered Lonely as a Cloud"\}} \\

Assistant & Asks the user for a featured image URL to include in the newsletter. \\

\rowcolor{gray!10}
User & Image link: \texttt{https://example.com/images/marigolds-with-grandchild.jpg}. \\

\rowcolor{blue!6}
Assistant (Tool Call) &
\texttt{ImageDescriber(image\_url="https://example.com/ images/marigolds-with -grandchild.jpg")} \\

\rowcolor{green!6}
Tool (Response) &
\texttt{\{"description": "A sunny garden scene ... 'Grandma's Sunshine Patch.'", "success": true\}} \\

Assistant & Proposes a family prompt question and formats the newsletter using \texttt{NewsletterFormatter}. \\

\rowcolor{blue!6}
Assistant (Tool Call) &
\texttt{NewsletterFormatter(headline="Planting Marigolds, Growing Memories", body\_text="Every spring, ...", image\_url="https://example.com/images/ marigolds-with-grandchild.jpg", quote="They flash upon that inward eye ...", attribution="--- William Wordsworth, from 'I Wandered Lonely as a Cloud'", question\_of\_week="What’s a garden, flower, or outdoor tradition that brings back warm childhood memories for you?")} \\

\rowcolor{green!6}
Tool (Response) &
Returns \texttt{html\_output} and \texttt{preview\_url} (omitted for brevity). \\

Assistant & Generates an optional audio narration for accessibility. \\

\rowcolor{blue!6}
Assistant (Tool Call) &
\texttt{AudioNarrationGenerator(text="Planting Marigolds, Growing Memories. Every spring, ...")} \\

\rowcolor{green!6}
Tool (Response) &
\texttt{\{"narration\_status":"success", "audio\_url":"https://example.com/audio/ narration/garden\_memory\_narrated.mp3"\}} \\

\rowcolor{orange!10}
Assistant (Final) &
\texttt{<answer>}The newsletter is generated (HTML preview available) and an audio narration link is provided for accessibility.\texttt{</answer>} \\
\bottomrule
\end{tabular}
\caption{An example of multi-turn interaction in our mock environment (user + tools) for the synthesized tool-use instance in Table~\ref{tab:tooluse_case}, illustrating user interaction, tool calls, and tool responses.}
\label{tab:interaction_case}
\end{table*}

\section{Contribution Summary}

Our method has clearly defined boundaries from prior work on data synthesis and environment simulation, mainly in the following aspects: (1) agentic training data synthesis, (2) stable environment simulation for agentic RL training, and (3) rubric-based rewards. These components have rarely been studied in previous studies.

Prior data synthesis work~\cite{wang2023self} has focused primarily on reasoning tasks, largely overlooking diverse tool-use scenarios. We synthesize task-specific tool ecosystems with deliberately underspecified instructions containing information gaps and user-private context. This necessitates multi-turn communication and long-horizon planning, promoting procedural generalization to new tools rather than memorizing interfaces or following scripted steps.

Existing work on environment simulation, such as ToolLLM~\cite{qin2023toolllm}, primarily targets benchmark construction. In RL, however, reproducibility is critical: if the simulator returns different tool responses for the same state and action, RL training becomes unstable. This issue is rarely addressed. We introduce task-level finite mappings that enforce consistent responses for identical tool calls within each task, yielding stable simulations.

We automatically derive execution-aligned, task-specific rubrics that cover subgoals, required interactions, and disallowed behaviors, and use them to build rubric-based RL rewards, an area rarely explored in agentic RL training.

These elements are integrated in \textsc{SynthAgent} as a closed loop rather than a loose collection. Each sample jointly specifies tools, instructions, hidden information, and evaluation criteria, yielding a reproducible agentic RL training recipe.

% \section{Example Appendix}
% \label{sec:appendix}

% This is an appendix.

\end{document}